# Words, Spaces and Generative AI: Layers of language in contemporary architecture

Anca-Simona Horvath

https://orcid.org/0000-0000-0000-0000

**Abstract**
Language can be considered a design material in architecture, and in the context of text-to-X generative AI models becoming a common tool for architectural practice, looking more closely at language is more important now than in the past. After describing some of the important developments in linguistics starting from Wittgenstein, and including the work of Chomsky, Lakoff, conceptual and generative metaphors as proposed by Schön, this chapter connects them to contemporary architectural design and generative text-to-X tools. The chapter builds on the idea that three main forms of language intertwine in architectural design done using generative AI, namely (I) discourse (or natural language which can contain professional terminology specific to our field), (II) programming languages (which are artificial languages sitting at the basis of all computational systems), and (III) annotations (as language elements attached to pieces of data). It concludes by outlining a research agenda for connecting generative metaphors to generative AI: (a) conducting corpus linguistics studies on architectural texts (using quantitative tools such as topic modelling, and qualitative tools such as discourse analysis); (b) bringing communication theory and information studies closer to architectural research and (c) taking into account that different (natural) languages come with different affordances meaning generative and conceptual metaphors differ in relation to this.

## 1. Introduction

Generative AI is now commonplace in architectural design practice. While many new models are being proposed at a speed that can be difficult to follow, text-to-X generative AI is among the most popular, especially due to it's ease of use and access. In this context, studying text as language, and its relationship to architectural design is more important today than in the past.

This chapter aims to unpack various dimensions of the relationship between language and architecture in the context of generative AI becoming prevalent and text-to-X generative models entering widespread use. It situates language as a design material in continuation of previous work that links linguistics and architecture. The relationship between architecture and language is considered well-documented in the architectural humanities, going as far back as since the inception of the field of modern linguistics proper at the end of the 18th century. Later, during the 1990's, ideas from Derrida's theories about deconstructivism entered architectural theory, and later practice with well-documented projects such as Peter Eisenman's work (Temple 2020) – which is considered unintelligible without the texts that accompany it.

In the influential book *Programming Architecture*, Paul Coates deals with programming languages as they can relate to architectural design and explains that there are two types of languages: natural languages, which have been developed during the last 100,000 years or so, and artificial languages – which have no native speakers. Programming languages are a type of artificial language (Paul Coates 2010). The previous waves of retooling the architectural profession made it clear that programming languages need to be given attention in architectural education, practice and research. Text-to-X generative AI demands we look more closely at natural language as well.

This chapter is organized as follows: in the next section we discuss some of the recent work on text-to-X generative AI models. After this, we give an overview of the historical relationship between architecture and linguistics followed by unpacking three layers of language that make up architectural design done with the use of generative AI, namely: architectural discourse (natural language and professional terms), programming languages (as artificial languages) and annotations as proposed in (Horvath and Pouliou 2024b). We then briefly discuss embodiment and language, as any current discussion on what language is and how it is used require an understanding of embodiment. Finally, in the last section, we outline possible future research directions for architecture in the context of text-to-X cultural production.

## 2. Text-to-X Generative Models: prompting architectural concepts, structures, elements

Recently, a large number of research studies describe the use of generative AI for architectural design education, or in creating a variety of architectural concepts or elements. For example, in a recently published paper, (Du et al. 2026), describe and share the code of a Vectorworks plug-in which allows text-to-BIM creation. The framework orchestrates multiple LLM agents to collaborate transforming textual user input into imperative code that invokes BIM authoring tool APIs, generating editable models with internal layouts, external envelopes, and semantic information directly in the software. Four specialized agents divide the labor: a Product Owner refining user inputs, an Architect developing structured building plans, a Programmer translating requirements into executable code, and a Reviewer checking output quality. Rule compliance rates approaching 99% were reported, though the authors acknowledge ongoing challenges with complex spatial layouts.

Bleker et al.'s *Text2Structure3D* (Bleker et al. 2026) trained a graph-based latent diffusion model on 30,000 synthetic bridge structures, each paired with ten automatically generated text descriptions ranging from brief (“a 60 metre span bridge”) to fully specified. Natural language prompts are converted into conditioning vectors via a pre-trained CLIP model, which steers a Diffusion Transformer to generate structural graphs — encoding node coordinates, force densities, and support conditions — close to static equilibrium. A residual force optimization step guarantees full equilibrium compliance in the final output.

In a recent paper, (Çelik 2026) analysed how multiple text-to-image generative models including Dal-E and Leonardo.ai perform when tasked with generating vernacular architecture across different contexts across Turkey, Japan, and Mexico. Using a standardised prompt system and a visual morphological analysis matrix, the study integrates expert evaluation with a so-called Analytic Hierarchy Process to assess form typology, material articulation, roof configuration, spatial organisation, openings, and symmetry. While models successfully approximate dominant formal and material cues, they consistently simplify the deeper spatial and cultural logics embedded in vernacular traditions, with Leonardo.Ai achieving the highest fidelity and DALL·E 3 second.

Pektaş and Sağlam's paper (Taşlı Pektaş and Sağlam 2025) deals with the trial-and-error way architects currently use text-to-image tools, arguing that the absence of structured frameworks produces outputs that are visually plausible but semantically thin. The paper proposes semiotics as an analytical method for crafting prompts that yield richer architectural outputs, testing its effects across DALL-E, Midjourney, and Stable Diffusion through two experiments examining how semiotic analysis and context modifiers affect output relevancy. Results confirm the method's effectiveness while exposing persistent problems of control, transparency, and training data quality — concluding that meaningful human-AI interaction in architecture requires a genuinely human-centred approach rather than optimised prompting alone.

Language-based instructions prescribing how a building should be or describing how it is like - has been commonplace for centuries in the forms of design briefs, legal specifications, discussions amongst professionals and with non-professionals, architectural commentary and theory. The examples above constitute a small part of the current fast growing landscape of developments in architectural research using text-to-X tools. But while much work is being conducted on the technical and directly practical aspects of the possibilities these tools allow, less research has focused on what this entails from a theoretical perspective.

## 3. Linguistics and its connection to architecture

There is a strand of work with a long history that has interpreted architecture *as* language – arguing that architecture itself can be understood as a visual and spatial language with its own form of grammatical structure. The most well-known example is Christopher Alexander's *A Pattern Language* which has had significant impact on both architecture and software engineering (Alexander 1977).

Peter Medway, a prominent researcher on language has spent some of his career trying to understand how language and architectural design intersect and stated in one of papers that: *"[...] architects don't build. All they do is configure signs, including the signs that constitute writing. Their configuring of signs — alphabetic, numeric, but also, of course, graphical (that is, drawings) — makes buildings go up."* (Medway 1996)

### 3.1. Language and concepts

In the early 1900's, philosopher Ludwig Wittgenstein developed a theory of language in the book *Tractatus Logico-philosophicus* (Ludwig Wittgenstein 1922). The book dealt with the relationship between words and things and asked how human beings manage to communicate ideas to one another. The proposition of Wittgenstein was that language triggers within us, humans, pictures / images of the things being communicated and words enable us to make pictures of facts. Communication is the process of swapping pictures between us. However, images and physical models are needed to communicate certain ideas because, on the whole, we are bad at creating images in the minds of others – which is the reason why communication often fails. For Wittgentein, drawing and creating physical models was important as an addition to spoken and written language, as they helped people better understand ideas communicated through words (Ludwig Wittgenstein 1953).

Various theories of what language is and how it used and have been developed over the last century and a half, with Wittgenstein's work being considered among the most important philosophical treaties of all times. Many of the theories developed afterwards build upon this prior work.

### 3.2. Generative metaphors

The idea of the generative metaphor was proposed in the late 70's and early 80's. The book *Metaphors We Live By* written by linguist George Lakoff and philosopher Mark Johnson (Lakoff and Johnson 1980) argues that linguistic concepts are cultural, context dependent and that they shape how humans experience reality. According to the two, the human conceptual system is largely metaphorical: we understand abstract ideas through concrete experiences, and metaphors are part of our brain's conceptual system. For example, when trying to understand an abstract concept such as time, we relate it to other – more concrete ideas. This as – *"the essence of the metaphor is understanding and experiencing one kind of thing in terms of another"*. Stating that "time flows/flies" or "time is money" or that "all the world is a stage" are examples of conceptual metaphors. These metaphors shape how we perceive, reason, and act in the world, often without our awareness, as– *"our conceptual system is largely metaphorical"* (Lakoff and Johnson 1980). The book became a foundational text in cognitive linguistics. In a text summarizing the work of Lakoff and Johnson, Kevin Clark argues that their work can be broken down into a few main claims, namely: (1) that any objective assumptions are problematic as all language and experience are embodied and thus contain subjectivity; the reason for this being that (2) the human mind is embodied; and because of this – (3) truth is relative to embodied understanding; (4) the way humans use categories in language is not easily defined, and the structure of these categories is non-classical; (5) conceptual systems consist of cognitive models; (6) thinking utilizes frames, metonymies, and prototypes; (7) metaphor is prevalent in human communication and primarily conceptual, similar to abstract thought; (8) image schemas structure our experiences. (Clark 2024)

One year earlier, in 1979, urban planner, and design researcher Donal Schön published a book chapter called *Generative Metaphor* in the book *Metaphor and Thought* (Ortony 1993). The chapter proposes that problem-solving in complex fields such as urban planning and policy depends also on how the problem is framed from the onset (Schön 1993). Schön argued that metaphors underlying policy discourse don't merely describe problems but generate the solution space we are able to think within: how a situation is framed determines what solutions appear sensible. Calling a slum "blight" implies it is necessary to remove it, while calling it a "natural community" invites its preservation. In this way - critical work happens before problem-solving, in problem-setting. As he put it, "problems are not given", but constructed by people trying to make sense of complex, troubling situations. Conflicting policy frames, he suggested, can be addressed through "frame restructuring"— surfacing the tacit metaphors that shape our perception of social reality.

Before Lakoff, Johnson and Schön, Noam Chomsky's work in linguistics, starting in the 1950s, argued that all human languages share structural similarities and that the capacity to learn language is an innate cognitive trait in humans. Chomsky's claim is that humans are born with an inherent universal grammar — the biological capacity for language which is not learned from experience, but hardwired into our biology and all human languages, however different on the surface, share deep structural principles. Chomsky made a distinction between competence (the unconscious knowledge of grammatical rules a speaker possesses) and performance (actual language use, which can be messy and error-prone). His most influential technical contribution was the *transformational generative grammar* — the idea that sentences have a deep structure that gets transformed into surface structure through rules. Therefore, language, for Chomsky, is fundamentally a property of the human mind, not of communication or social interaction. Apart from this – Chomsky's argument also alludes to the idea that language is embodied, meaning there is a direct relationship between the human body and the ability for and of language.

Chomsky's theory of language was an inspiration to the development of L-systems by the Hungarian botanist Lindenmayer in 1968 (Lindenmayer 1968b, 1968a). Lindenmayer was studying the growth patterns of algae and filamentous organisms, and wanted a mathematical formulation to describe how cells divide and develop in parallel. In trying to achieve this, he developed L-systems (Lindenmayer systems) - formal grammars for modelling the growth of biological organisms — plants, branching structures, fractals. These work by applying rewriting rules repeatedly to strings of symbols, generating complex structures from simple rules. Lindenmayer borrowed the conceptual apparatus of formal grammars — alphabets, production rules, strings — directly from Chomsky's hierarchy, established in the 1950s. But he modified it: where Chomsky's grammars rewrite one symbol at a time sequentially, L-systems rewrite all symbols simultaneously. L-systems are a type of formal grammar in the Chomskian sense, sitting within his hierarchy of generative grammars. However, L-systems differ in that all rules apply simultaneously at each step, mimicking parallel biological growth. Applications of L-systems proved broad and includes fields such as computer graphics — where L-systems are used in procedural generation of realistic plants and trees (Pixar, game engines); developmental biology — used for modelling morphogenesis and organ formation; but also architecture — where L-systems have been used in generative design and parametric form-finding and urban planning — used for simulating street network growth.

Chomsky was the PhD supervisor of Lakoff, and in the initial stages of his career, Lakoff worked on Chomsky's theories on language. Over time however, the two diverge significantly with Lakoff arguing that language is an inherently social activity, fluid, dependent on cultural context, and Chomsky arguing that languages are rule-based and hardwired. Both these theories are important in architecture in the current context, as we will detailed later.

Another important – although slightly less-well known theory on language was developed in the 1970s by psychologist Eleonor Rosch called *prototype theory* which is a model in cognitive psychology and linguistics showing how humans categorize objects and concepts. In her work, Rosch looked at how people think about concepts (through language, and also how language is transformed into images – building on Wittgenstein's work) and established that not all concepts are equal, and that they function based on hierarchies and categorization. For example, Rosch showed that when thinking about the following categories "furniture" -> "chair" -> "armchair" –people find it easier to put an image to the concept of "chair" or "armchair" than to imagine the category of furniture, which is broader (Varela, Thompson, and Rosch 2017; Mervis and Rosch 1981; Rosch 1975a, 1975b; Rosch et al. 1976). In general, categories do not have strict boundaries, and some concepts have boundaries that are looser than others – in other words categories have fuzzy edges. In an article detailing the work of Rosch and her colleagues, James Hamptom explains that: "*The central insight of prototype theory is that word meanings, and the conceptual classes that the* *words name, are distinguished one from another not in terms of an explicit definition, but in terms of similarity to a generic or best example. The concept red is the class of colors that are centered around a particular point on the spectrum that everyone tends to agree is the prototype red.*" (James A. Hampton 2006). People can learn new categories based around prototypes.

All these theories are relevant in the context of generative AI text-to-X architectural production. Chomsky's work has proven to be more relevant to computer science and programming languages and in emulating geometric natural patterns. Lakoff's work remains among the most important in understanding human languages and human cognition – so it is relevant in terms of understanding how people communicate when aiming to explain design decisions or constraints. Rosch's ideas translate directly into how annotations of images work in people's minds, and could be further researched in terms of how to design text-to-image and text-to-3D model systems which can have direct applications to design fields.

## 4. Layers of language in architectural design

(Horvath and Pouliou 2024b), shows that three layers of language intertwine when working with generative AI tools such as text-to-text or text-to-image tools, namely: natural language (that can include professional terminology), programming languages (which are artificial languages that sit at the basis of all computing technologies), and annotations (which are words that describes images or other pieces of data). We build upon this categorization here, and unpack each of them below, as they connect to the theories in linguistics presented in the previous section.

### 4.1. Discourse

A portion of architectural knowledge is recorded and distributed in textual form. These can include legal documents prescribing what and where can be built, design briefs, abstracts describing architectural concepts. This discourse can take the shape of written or spoken language.

#### 4.1.1. Writing architecture [types of written texts about architecture]

Among the more important work that has looked at written texts about architecture from different angles is *Words and Buildings: A vocabulary of modern architecture* (Forty 2000). In this work, Forty starts by describing architecture as a three-part system formed by: (a) the building; (b) its image (images prescribing how buildings should be, or images describing existing buildings) and (c) its accompanying critical discourse. The book goes on to unpack the vocabulary of modernist architecture showing how specific words such as grid, form, space, type acquire particular meanings within modern architecture demonstrating that architects use language not merely to communicate ideas but to construct the architectural concepts themselves. By revealing contradictions and shifts in architectural vocabulary, Forty challenges the assumption that architecture can be understood independently of discourse. Language, for Forty, is part of architecture's material and intellectual production (Forty 2000).

In the book *The Words Between the Spaces*, architect Thomas Markus and linguist Deborah Cameron, discuss precisely the relationship between architecture and language (Cameron and Markus 2003). The book examines how language participates in the production, interpretation, and use of buildings. The two authors combine architectural analysis with discourse analysis to show that buildings are surrounded by texts—briefs, regulations, criticism, histories, classifications, and promotional language. They argue that these texts do not simply describe architecture but actively shape its meaning, organization, and social effects. Through cases including the Scottish Parliament, Berlin's Reichstag, and Auschwitz, they examine how language constructs classification, power, value, heritage, and architectural images (Cameron and Markus 2003). The book therefore positions language as an essential component of architecture.

More recently, the article *How we talk(ed) About It: Ways of speaking about computational architecture*, (Horvath 2022), presents a corpus linguistics analysis (an analysis on a large corpus of texts) of architectural texts formed by: the abstracts from winning projects and honourable mentions of the eVolo skyscraper competition between 2006 and 2022, together with the texts of the articles published in Architectural Design between 2005 and 2022. Through a quantitative and then qualitative analysis of more than 4,5 million words, the article unpacks how various words are more or less popular at different times in recent architectural texts, showing, for example, that the word 'human', and words connected to it (such as 'people', 'social', or 'person') have periods when they are more or less popular over the years. The article shows how the language used to talk about and theorize computational design in architecture makes use of specific terminology, and how this language changes over time reflecting changes in the field and in research and practice more broadly. The work also highlights how concepts coming from adjacent scientific fields enter the discourse of architecture – and how they are popular at different times. While the title references what appears to be "spoken" language, the article focuses on written texts about architecture.

In a study from 2023, (Yazici and Durmus Ozturk 2023) study a corpus of texts written by Rem Koolhaas on architecture and urban planning. Using the AntConc corpus analysis toolkit on Koolhaas's articles from 1977 to 2014, the study transforms his texts to words and lexical bundles, focusing on collocation — words that habitually appear together. The paper examines the discourse codes embedded in Koolhaas's language and the metaphors structuring his thinking about architecture and urbanism. The authors argue that a corpus-based model applied to architectural texts has the potential to open new areas of knowledge for architectural discourse, and that deciphering textual codes contributes to developing new analytical models. While corpus linguistics studies – using natural language processing tools widely available today - on architecture have the potential to reveal important ways in which knowledge in our field is produced, they remain rare.

#### 4.1.2. Talking about it [the use of spoken language in spatial design]

In an article from 2011 titled *Messy Talk and Clean Technology* (Dossick and Neff 2011), structural engineer Carrie Dossick together with communication scholar Gina Neff present an ethnographic study of how digital transformation happens in practice in the architecture, engineering and construction industry, around the use of Building Information Modelling (BIM). They show how the unstructured communication sessions between the different specialists who talk over printed floorplans and sections during meetings, or during sessions when the specialists present their work and ideas to clients (i.e. non-specialists). The article shows that important decisions are made during these sessions where "messy talk" takes place, and that BIM (the clean technology) does not capture such communication.

Inspired by this work, (Horvath et al. 2021) created a VR interactive experience that allowed non-professional users to access the BIM of a large-scale architectural project by interacting with every object in the scene, being able to write comments about it as a means to open the BIM, and facilitate communication.

While some research has focused on the digital transformation of the AEC sector – from the point of view of communication and information science, this strand of work remains small, and is rarely taken up, cited and built upon in architectural research (Vite et al. 2021; Pouliou, Palamas, and Horvath 2024; Horvath and Pouliou 2024a; Gardner 2022).

### 4.2.Programming languages

Apart from written and spoken language, another layer of language that makes up generative text-to-X tools are programming languages. These languages are artificial – as they have no native speakers - and they form the basis of all computational systems. Paul Coates' book *Programming Architecture* deals specifically with programming languages linking them to computational design within architecture. His theoretical anchoring for the book is Chomsky's work on the universal grammar (Paul Coates 2010).

Within computer science, the book *Understanding Computers and Cognition* written by computer scientist Terry Winograd and philosopher Fernando Flores (Winograd and Flores 1986) was considered foundational in explaining the evolution of programming languages with their connection to natural language. The book is a s a landmark critique of the rationalist tradition in AI and interface design (which persists today, 40 years after the books publication), drawing on Heidegger, Maturana, and speech act theory (the philosophical claim that language does not merely describe the world — it does things in the world). Their argument about programming languages is that conventional languages are designed on the assumption that computation is about representing and manipulating a pre-given world. Winograd and Flores reject this, arguing that language — including programming language — does not describe reality but constitutes it through action. Drawing on Austin and Searle's speech act theory, they reframe programs as commitments within networks of human conversation. A program is less like a map of the world and more like a promise — it creates obligations, opens possibilities, and is embedded in a social context that no formal specification can fully capture. It remains one of the most philosophically serious attempts to think programming languages not as neutral tools but as shaped by — and shaping — how we understand what computers are *for*.

Programming languages are in a sense easier to deconstruct than natural languages as they have specific structures and syntaxes and are to be used in a precise way (they are more clean than the messy communication between people). Their purpose is to translate commands (that are readable by humans) into strings of 0's and 1's readable by machines. However, as (Winograd and Flores 1986), show, they remain *tools for communication*, which are not neutral, they open certain possibilities, and close others – in other words, they have certain affordances.

### 4.3.Annotations

Finally, a third layer of language involved in text-to-X AI tools are annotations. The ways in which annotations on images were initially created in the technology that fuelled the current AI breakthrough has been well-described in many places, including the book *Atlas of AI: Power, Politics and the Planetary Costs of Artificial Intelligence*, by Kate Crawford (Crawford 2021). In asking what artificial intelligence actually is, Crawford embarked on a journey to trace where data centres are built, how big data is mined and processed so that it can be used to train advanced AI models. On this journey, she finds that images used to train text-to-image generative AI are annotated in so-called annotation farms placed in the developing world. The most consequential dataset in the history of AI is called ImageNET (a large-scale visual database containing over 14 million hand-annotated images organised into more than 20,000 categories), was modelled based on a word corpus linguistics, used to train text-to-text generative AI and recommender systems, called WordNet. WordNet is a lexical database of English developed at Princeton University by George Miller and colleagues, first described in the paper *WordNet: A Lexical Database for English* (Miller 1995). WordNet organized approximately 170,000 words into around 117,000 synsets — groups of cognitive synonyms — linked by semantic relations including hypernymy (dog → animal), meronymy (wheel → car), and antonymy. This structure is similar to how Eleonor Rosch described concepts in language – with concepts being hierarchical.

In AI training, WordNet has been used in two main ways. First, it served as a labelling scaffold: ImageNet, the dataset that triggered the deep learning revolution, used WordNet's noun hierarchy directly to organize its 14 million images into categories — every image label corresponds to a WordNet synset. WordNet's structure, with all its embedded assumptions about what things are and how they relate, was baked into the models trained on ImageNet, including early versions of systems now ubiquitous in visual recognition. Second, WordNet has been used for semantic similarity tasks in natural language processing, helping models learn that *sofa* and *couch* are closer in meaning than *sofa* and *apple*. The deeper consequence, as scholars like Crawford have noted, is that WordNet's particular way of classification – sits at the basis of the current AI models which are in broad use today. WordNet was initially constructed from texts as diverse as legal documents, food recipes, and dictionaries. In 2009, Stanford researcher Feifei Li, published ImageNet with a structure based on that of WordNet, but instead of using all of wordNet's synsets, the researchers chose on only map the nouns, as it was considered easier to map a noun to an image.

According to Crawford, building on work from Lakoff, Rosch and others, nouns sit on an axis between more concrete and more abstract. For example, an apple can be considered a more concrete, nouny noun, than nouns such as "health" or "debt", and when it comes to categorizing people this becomes highly problematic: "*in the case of the 21841 categories that were originally in the ImageNet hierarchy, noun classes such as 'apple' or 'apple butter' seem reasonably uncontroversial, but not all nouns are created equal. To borrow an idea from linguist George Lakoff, the concept of an 'apple' is a more nouny noun than the concept of 'light' which in turn is a more nouny noun than a concept such as 'health'. Nouns occupy various places on an axis from the concrete to the abstract, from the descriptive to the judgemental.*" (Crawford 2021). The argument the book makes is that the categorizations include high levels of bias. Crawford and Paglen's article *Excavating AI* (Crawford and Paglen 2021) exposed how ImageNet classified people under categories including offensive slurs and psychiatric diagnoses, revealing that the dataset's apparently neutral organisational logic encoded deep social prejudices. Princeton removed over half a million person-category images in 2019 following public pressure, though the models already trained on that data remain in deployment.

This is relevant to architectural theory and practice as it makes evident that understanding language, language in use, communication theory together with prototype theory, and how conceptual metaphors are used in design practice should inform text-based annotations, for text-to-X architectural production.

## 5. The nature of embodiment in language

Most theories on language agree that natural languages are embodied, and relate to how humans or other species experience the world through their bodies and sensory apparatus. Wittgenstein's understanding of language takes into account that no two people experience the world in the same way, and that these experiences shape how they create images from linguistic concepts. While Chomsky's work states that languages have inherent rules, he also argues that humans have an innate, biological ability for language. The subsequent work of Lakoff and Rosch define language as embodied and connected to how the human body holds its perceptual systems which has been demonstrated empirically and is widely accepted neuroscience and cognitive psychology today (Lisa Feldman Barret 2017).

This is significant because computational systems have different forms of embodiment from humans (they sit in non-organic physical machines), and so communication for and with these systems cannot be the same as communication between humans. Nevertheless, this is rarely acknowledged. According to (Alexander Bie and Mark Grimshaw-Aagaard 2026), AI systems have so-called inherited phenomenology, meaning they are trained on data generated by humans, which embed classifications and concepts based on human embodiment. However – there is a complex translation process that needs to take place to translate human embodiment into machine embodiment. This becomes more important when dealing with robotic systems. Therefore, we can see text-to-X generative AI as a process of communication – and this should draw on research and knowledge from the field of communication science, as well as information science.

## 6. From generative metaphors to generative AI: Drafting future research directions for language as a design material in architecture

As shown in these previous sections, the relationship between language and architecture is complex, and long-standing. While a significant portion of work over the last centuries has tried to understand architecture *as* language and putting it in relationship with linguistics, the relationship between architecture *and* language has been less researched. As previously discussed, multiple layers of language come together in architectural design that makes use of text-to-X generative tools - shown in Table 1.1.[1].

| Layer of language | Characteristics |
|---|---|
| Spoken **discourse** / talk [professionals talking about architecture, i.e. in conferences, interviews etc] | Occurs using **natural language** (has native speakers), but the language includes professional terminology; specific concepts are used in ways that might differ from how they are used in common language. Examples include: "urban tissue", or "digital morphogenesis". |
| Written **discourse** – descriptions / abstracts explaining concepts | Using **natural language** (has native speakers), but includes professional terminology. More formal than |

[1] Table 1.1. – three layers of language that come together in architectural design done with the help of generative text-to-X tools.

| | |
|---|---|
| and aspects of thought processes / architectural theory and research | spoken discourse. Specific concepts are used in ways that might differ from how they are used in common language. |
| **Programming language** | **Artificial languages** (no native speakers); evolved from a series of 0's and 1's (representing electrical current passing or not in a circuit) to modern interpreted programming languages which are easier to read by humans – meaning they are closer to natural languages. |
| **Annotations** | **Hybrid between artificial and natural language** (simplified natural language); representations of images |

There are also a series of theories about language and linguistics, demonstrating what language is, and how it shapes communication between people as described in the previous section, these are summarized in Table 1.2 [2].

| Theory of language | Main argument |
|---|---|
| Wittgenstein (early 1900) | Human communication functions with people creating images in their minds from the words the listen to. Meaning / concepts communicated through words are translated into mental images. |
| Chomsky (~1950-) | Humans have innate ability to learn languages. Language have syntaxes that order them: a universal grammar exists which is a the basis of human languages. |
| Lakoff&Johnson (~1970-) | Humans use metaphors to make sense of the world. complex concepts such as time, debt of health are understood through concepts. The human perceptive system is metaphorical. |
| Rosch (~1970-) | Prototype theory – humans organize the world through prototypes — idealized mental representations —meaning categories are structured around best examples. The human conceptual system categorizes concepts, i.e. furniture->chair->armchair. It is easier to put a mental image to a middle category (i.e. chair), than to a top category (i.e. furniture). |

Computational tools in general are based on programming languages which are forms of artificial languages. Language can be understood as one of three parts that forms architecture, alongside images and the built environment itself. However current developments in AI further complicate the relationship between language and architecture given the rapid development and increase in use of text-to-X tools. While much work is going into developing multi-modal AI systems that take into account different forms of architectural data, some researchers with the field of human-computer interaction argue that text-to-X tools will bring a "death of the interface" (Phillip Gallagher 2024). The reason for text-to-X generative AI's fast adoption is its inherent ease of use. People no longer need to learn a new interface in order to generate various forms of media. Moreover, they can easily transition from one tool to another with little friction, as the interfaces of generative AIs are all similar and based on natural language. However, if we are to build up on the work of Wittgenstein, Lakoff, Rosch – drawings and models are forms of communication that are more precise than spoken language, although harder to produce. In practice – this might mean than graphical and volumetric representations for architectural design will remain important means of communicating precise ideas which are difficult to explain through natural language.

[2] Table 1.2. – selected theories of language and linguistics that have applications in architectural design done with the help of generative text-to-X tools.

Given these developments, architectural research will need to engage more specifically with and build upon previous research on language, communication and information science in order to unpack the practical and theoretical implications of generative text-to-X tools. This future research can take multiple directions as outlined below.

On the one hand, conducting corpus linguistic studies using quantitative (i.e. natural language processing, topic modelling, sentiment analysis) and qualitative research methods (i.e. discourse analysis) can illuminate long-term trends in how specific concepts (which become generative metaphors/concepts) are used in our field, and how they are employed in making (sense of) the physical world.

Studying communication about the process of designing buildings using computing technologies, as similar to the work of (Dossick and Neff 2011; Vite et al. 2021) is also important in understanding spatial cognition from the point of view and from within our field. In this case, using research methods from human-computer interaction and human-robot interaction can illuminate how designers interact with the new graphic-less interfaces, and can be later translated into knowledge on how generative AI tools and the datasets that feed them should be tailored for our field, based on knowledge developed in the previous step which unpack how conceptual metaphors structure thinking and later action.

Third, natural languages themselves have different affordances, and speaking a certain language will determine people's experience of the world, as work within cognitive linguistics and neuroscience show (Lisa Feldman Barret 2017). This means that both corpus linguistics studies looking at how different languages are used within architecture (to gain an understanding of conceptual metaphors that are language specific), as well empirical studies where speakers of different languages engage with generative text-to-X tools are important in understanding how professionals and non-professionals use language when proposing solution or building architectural space.